\documentclass{article}

\usepackage{microtype}
\usepackage{graphicx}
\usepackage{subcaption}
\usepackage{booktabs} %

\usepackage[breaklinks]{hyperref}

\usepackage[accepted]{icml2026}

\usepackage{amsmath}
\usepackage{amssymb}
\usepackage{mathtools}
\usepackage{amsthm}

\usepackage[capitalize,noabbrev]{cleveref}

\theoremstyle{plain}

\theoremstyle{definition}

\theoremstyle{remark}

\usepackage[textsize=tiny]{todonotes}

\usepackage{colortbl}
\usepackage{array}
\usepackage{multirow}
\usepackage{enumitem}
\usepackage{tabularx,ragged2e}
\usepackage{float}
\usepackage{inconsolata} %

\usepackage{microtype}

\newcolumntype{L}[1]{>{\raggedright\arraybackslash}p{#1}}

\newcommand{\RR}{\mathbb{R}}

\DeclareMathOperator{\topk}{TopK}

\DeclareMathOperator{\enc}{enc}
\DeclareMathOperator{\dec}{dec}
\DeclareMathOperator{\encoder}{encoder}
\DeclareMathOperator{\decoder}{decoder}
\DeclareMathOperator{\relu}{ReLU}

\icmltitlerunning{Position: Use Sparse Autoencoders to Discover Unknowns}

\begin{document}

\twocolumn[
  \icmltitle{Position: Use Sparse Autoencoders to Discover Unknowns}

  \icmlsetsymbol{equal}{*}

  \begin{icmlauthorlist}
    \icmlauthor{Kenny Peng}{equal,cornell}
    \icmlauthor{Rajiv Movva}{equal,berkeley}
    \icmlauthor{Jon Kleinberg}{cornell}
    \icmlauthor{Emma Pierson}{berkeley}
    \icmlauthor{Nikhil Garg}{cornell}
  \end{icmlauthorlist}

  \icmlaffiliation{cornell}{Cornell University}
  \icmlaffiliation{berkeley}{UC Berkeley}

  \icmlcorrespondingauthor{Kenny Peng}{kennypeng@cs.cornell.edu}
  \icmlcorrespondingauthor{Rajiv Movva}{rmovva@berkeley.edu}

  \icmlkeywords{Machine Learning, ICML}

  \vskip 0.3in
]

\printAffiliationsAndNotice{\icmlEqualContribution}

\begin{abstract}
While sparse autoencoders (SAEs) have generated significant excitement, a series of negative results have added to skepticism about their usefulness. Here, we establish a conceptual distinction that reconciles competing narratives surrounding SAEs. We argue that even if SAEs may be less effective for \textit{acting on known concepts}, SAEs are especially powerful tools for \textit{discovering unknown concepts}. This distinction separates existing negative results from positive results, and suggests several classes of SAE applications. Specifically, we outline use cases for SAEs in (i) ML interpretability, explainability, fairness, auditing, and safety, and (ii) social and health sciences.
\end{abstract}

\section{Introduction}

\setcounter{footnote}{0}

Sparse autoencoders (SAEs) have been a popular topic in interpretability research, showing impressive capabilities for identifying interpretable directions in the text representations underlying language models \citep{cunningham_sparse_2023, templeton2024scaling}. For example, an Anthropic paper found a ``Golden Gate Bridge'' direction, which could be manipulated to make a chatbot that would always incorporate the Golden Gate Bridge into responses \citep{anthropic2024goldengate}.

However, two recent papers show that SAEs fail to outperform simple baselines in large-scale evaluations on concept detection (probing) and model steering \citep{kantamneni2025sparse, wu2025axbench}. These results have led to pessimism about the usefulness of SAEs. For example, in response to this research, the mechanistic interpretability team at Google DeepMind announced that they would deprioritize research into SAEs \citep{deepmind2024sae}. Nonetheless, there continues to be optimism about new applications of SAEs, including in hypothesis generation \citep{movva_sparse_2025} and in the ``biology'' of LLMs \citep{lindsey2025biology}.

How can we square continued interest in SAEs with thorough evaluations demonstrating negative results? Are new attempts to use SAEs misguided? Or is there something missing in our understanding of the negative results? This position paper reconciles conflicting narratives surrounding SAEs by making a conceptual distinction. \textbf{Our position is that SAEs---even if less effective for \textit{acting on known concepts}---are powerful tools for \textit{discovering unknown concepts}.}

Consider the tasks where \textit{negative} results have been shown. Concept detection involves detecting a known, prespecified concept (``Does this text mention dogs?''). Model steering involves steering a model to exhibit a specified concept (``Make outputs less sycophantic.''). 
Another negative result involves concept unlearning \citep{farrell2024applying} (``Unlearn knowledge about concepts related to biosecurity.''). 
In these tasks, concepts are \textit{inputs}---known beforehand.

Now consider the tasks where \textit{positive} results have been shown. Hypothesis generation involves finding concepts that predict a target variable (``What concepts predict engagement of news headlines?''). Biology of LLMs involves finding concepts that LLMs represent when generating text (``What concepts does an LLM represent when doing addition?''). In both tasks, concepts are \textit{outputs}---unknown beforehand.
\footnote{When we say ``unknown'' concept, we mean that the researcher or practitioner cannot specify the particular concept that answers their question. This concept (once revealed to the researcher) may be well understood. For example, a researcher may be very familiar with the concept of ``surprise,'' but did not know beforehand that this concept predicted an outcome of interest (e.g., number of clicks on a headlines). Furthermore, the concept also may be stored in language model representations; therefore, the primary task is to uncover the right concepts—a task we argue SAEs are well-equipped to do.}

So even given results showing that SAEs underperform baselines when acting on known concepts, SAEs remain promising and underexplored tools for discovering unknown concepts. By enumerating concepts in an unsupervised manner, SAEs allow for the discovery of concepts that fit desired criteria. We outline how SAEs---as a tool for generating unknown concepts---can be used to advance research in (i) ML interpretability, explainability, fairness, auditing, and safety, and (ii) discovery in the social and health sciences. For example, in ML fairness and auditing, researchers can use SAEs to discover previously unknown concepts that bias model outputs. In the health sciences, researchers can use SAEs to discover previously unknown concepts that predict health outcomes, or to discover spurious correlations in existing predictive models.

(On the other hand, we do not intend to suggest that SAEs do not also have potential in the \textit{known concepts} regime. There are increasingly positive results in this setting compared to baselines \citep{cywinski2025saeuron, arad2025saes}, especially when considering cost and efficiency \citep{nguyen2025deploying}. Our paper focuses on establishing how SAEs are particularly useful in the \textit{unknown concepts} regime, and that negative results in the \textit{known concepts} regime need not dissuade researchers from exploring the use of SAEs.)

\paragraph{Paper structure.} Section 2 serves as a primer on SAEs (which can be readily skipped by readers familiar with SAEs). Section 3 shows that negative SAE results pertain to tasks that act on known concepts. Section 4 surveys recent positive results, showing that these papers use SAEs to discover unknown concepts. Section 5 then explores use cases for SAEs in different research areas.

\section{An SAE Primer}
\label{sec:primer}

We offer a brief primer on the SAE architecture, their history, and why and how they are now being used to interpret language models. (Readers familiar with SAEs may skip to Section \ref{sec:negative_results}.)

\paragraph{Early work on autoencoders.} Autoencoders are unsupervised neural networks that learn to reconstruct high-dimensional inputs via a series of learned transformations. 
For a $D$-dimensional input $\mathbf{x}$, an autoencoder computes:
\begin{align}
    \mathbf{z} = \encoder(\mathbf{x}), \\
    \hat{\mathbf{x}} = \decoder(\mathbf{z}),
\end{align}
where $\encoder(\cdot), \decoder(\cdot)$ are arbitrary neural networks, $\mathbf{z}$ is the \textit{latent} feature representation, and $\hat{\mathbf{x}}$ is the \textit{reconstruction}. The autoencoder is trained with a mean squared error reconstruction loss, 
\begin{align}
    \mathcal{L} = ||\hat{\mathbf{x}} - \mathbf{x}||_2^2.
\end{align}
One classic application of autoencoders is compression: by restricting the latent representation $\mathbf{z}$ to a dimension size $M \ll D$, the autoencoder learns a compressed representation in $\mathbf{z}$ which can be used to approximate $\mathbf{x}$ \citep{hinton_reducing_2006}.
In this setting, $\mathbf{z}$ functions similarly to an $M$-dimensional principal component analysis of $\mathbf{x}$,\footnote{If the encoder and decoder are linear, PCA minimizes the reconstruction loss \citep{baldi_neural_1989}.} in that we wish to explain as much variance as possible in the distribution of $\mathbf{x}$ using only $M$ dimensions.

\paragraph{Sparse autoencoders.} Sparse autoencoders (SAEs) perform the same reconstruction task, but leverage a different intuition. 
In an SAE, $M$ can be \textit{larger} than $D$, but each individual $\mathbf{z}$ is forced to be sparse---that is, only a small number of its dimensions can be nonzero. 
This design is motivated by the idea that while an entire dataset (e.g., all text on the Internet, or all images in ImageNet) may span many possible concepts, a single datapoint (a sentence or image) often contains very few.
Empirically, enforcing this structure often produces latents in $\mathbf{z}$ that capture atomic concepts.
For example, early work on SAEs trains directly on input images $\mathbf{x}$ (e.g., from MNIST), and $\mathbf{z}$ contains interpretable features like edges \citep{coates_importance_2011, makhzani_ksparse_2014}.

To improve clarity, we define \textit{features} and \textit{concepts}:
\begin{itemize}
    \item A \textbf{feature} is one of many numerical values used to represent an input. In a neural network, a feature is a single \textbf{dimension} of a layer's output vector.
    In an SAE specifically, a feature is an \textbf{activation} computed by a single \textbf{latent neuron}. 
    We use the terms feature, activation, neuron, and dimension interchangeably depending on context.
    \item A \textbf{concept} is a qualitative characteristic that may or may not be present in a given input. Here, we operationalize concepts via natural language descriptions.
    \item An \textbf{interpretable feature}, then, is a feature whose values correspond to the presence or absence of a single concept.
\end{itemize}

\begin{table*}
\includegraphics[width=\linewidth]{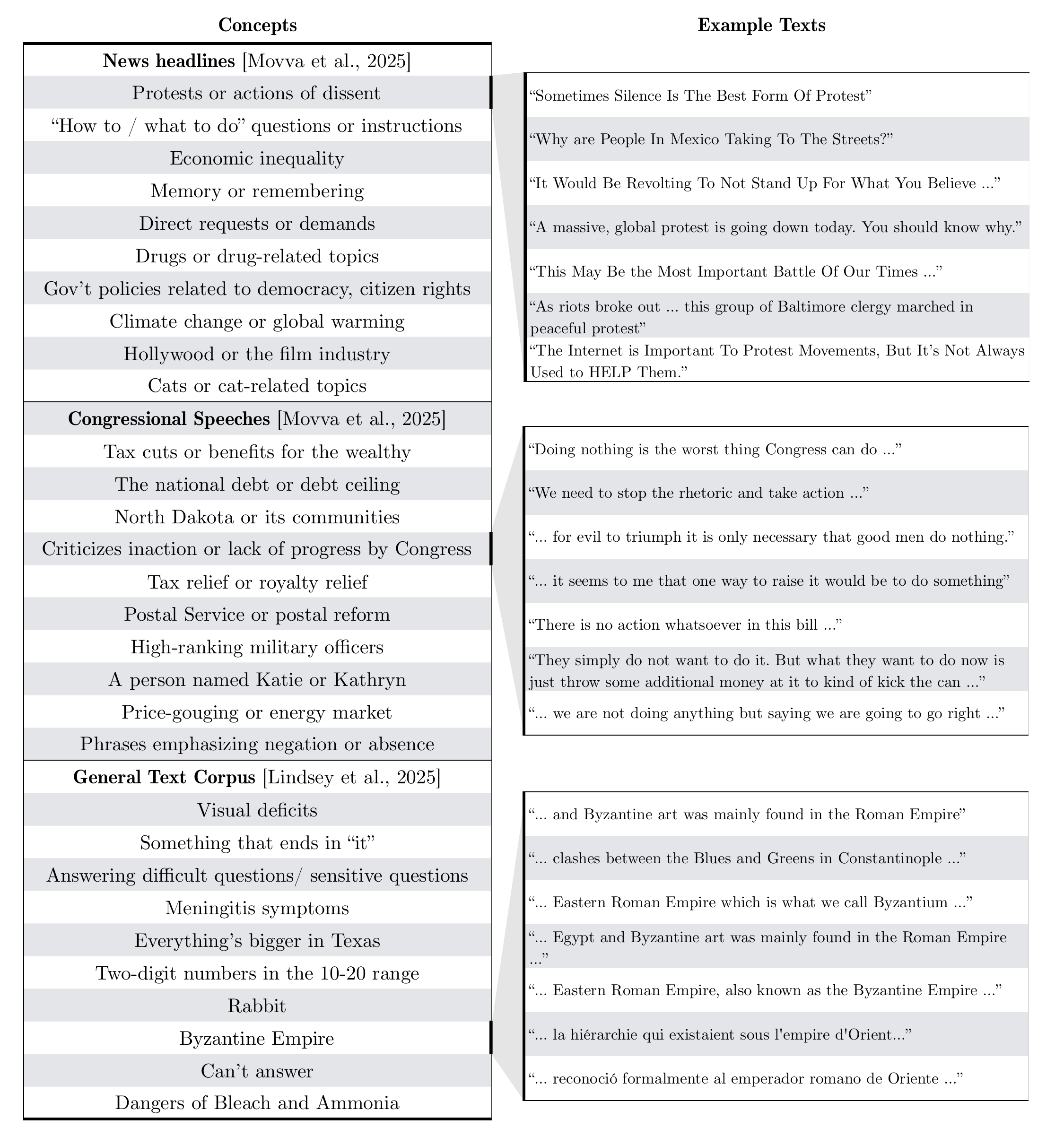}
    \caption{SAE latent features mapped to natural language concepts via autointerpretation, and texts that activate them \citep{movva_sparse_2025, lindsey2025biology}. \textbf{Left:} Examples of concepts learned from SAEs trained on different datasets; \textbf{Right:} Examples of texts that activate the corresponding SAE feature. Concepts interpretably describe the underlying data distribution of texts.}
    \label{tab:sae-examples}
\end{table*}

\paragraph{Mathematical formulation of SAEs.} One formulation of the sparse autoencoder follows the usual autoencoder forward pass, but adds an $L_1$ penalty on $\mathbf{z}$ to the loss function \citep{coates_importance_2011}:
\begin{align}
    \mathcal{L} = ||\mathbf{\hat{x}} - \mathbf{x}||_2^2 + \lambda ||\mathbf{z}||_1.
\end{align}
A larger $\lambda$ encourages more zero elements in $\mathbf{z}$. 
Another approach removes the $L_1$ penalty, but explicitly applies a $\topk$ function to the encoder that zeroes out all but the $K$ largest activations in $\mathbf{z}$ \citep{makhzani_ksparse_2014}, where $K \ll M$.
The full forward pass is given by:
\begin{align*}
    \mathbf{z} &= \relu(\topk(W_{\enc}(\mathbf{x} - \mathbf{b}_{\mathrm{pre}}) + \mathbf{b}_{\enc})),\\
    \hat{\mathbf{x}} &= W_{\dec}\mathbf{z} + \mathbf{b}_{\dec},
\end{align*}
where $\mathbf{b}_{\mathrm{pre}} \in \RR^D, W_{\enc}\in \RR^{M\times D}, \mathbf{b}_{\enc}\in \RR^{M}, W_{\dec}\in \RR^{D\times M}, \mathbf{b}_{\dec}\in \RR^{D}$.

Top-$K$ SAEs with a single-layer encoder and decoder (as above) have emerged as a common architecture in recent work, with slight variations to mitigate issues like dead neurons and feature absorption \citep{gao_scaling_2024, bussmann_learning_2025}. 
Some work has replaced SAEs with sparse \textit{transcoders}, which use layer $\ell_i$ to construct the output of a later layer $\ell_j$ \citep{paulo_transcoders_2025, lindsey2025biology}. 
For convenience, \textbf{we refer to all of these closely-related sparse coding methods under the ``SAE'' umbrella}, while noting that the specific optimal architecture is likely to shift.

\paragraph{Applying SAEs to interpret language models.} The recent wave of SAE research aims to interpret the representations learned by large language models. 
The motivation for this line of work is to understand the units and computations an LM uses to map inputs to outputs.
Before SAEs, a plethora of works over the last decade on \textit{probing} language models have shown that LM token representations contain rich semantic information \citep{belinkov_probing_2022}.
Concepts like a word's part-of-speech or pronoun co-references are a linear transformation away from the word's representation \citep{liu_linguistic_2019}.
Given this richness, a natural question is whether we can identify all of the concepts that a language model encodes. 
Unfortunately, neurons in a language model tend to be \textit{polysemantic}: they encode many concepts at once \citep{elhage_toy_2022}, so it is difficult to extract an LM's concepts by studying its neurons.

This convergence of findings---that language model representations encode numerous valuable concepts, but studying individual neurons does not reveal them---explains recent excitement for sparse autoencoders.
Unlike LM neurons, SAEs produce \textit{monosemantic} neurons that can be explained by a single concept \citep{cunningham_sparse_2023, bricken2023towards}. 
SAEs are trained on an LM's representations $\mathbf{x}$ of individual tokens, resulting in latent representations $\mathbf{z}$.
To interpret a particular feature dimension $i$ in $\mathbf{z}$, we can examine tokens that produce large values of $\mathbf{z}[i]$.
Initial work reports that after training on the representations from a small one-layer LM, the SAE features have succinct meanings like ``\textit{Arabic text}'' or ``\textit{citations in scientific papers}'' \citep{cunningham_sparse_2023, bricken2023towards}.
Follow-up work demonstrates that SAEs scale to state-of-the-art LLMs \citep{templeton2024scaling, gao_scaling_2024}, and that they can be used to interpret embeddings of text chunks rather than just individual tokens \citep{oneill_disentangling_2024}.
In Table \ref{tab:sae-examples}, we provide examples of concepts learned on both specific text datasets (news headlines and Congressional speeches) as well as general text corpora. SAEs may also be applied to other models, such as vision or biological models 
\citep{simon_interplm_2025a,brixi_genome_2025,adams_mechanistic_2025}.

\paragraph{Automatically interpreting features with language models.} While SAEs produce features in $\mathbf{z}$ that are theoretically interpretable, generating a mapping from features to concepts is a separate challenge. 
There are too many features to explain manually, so prior work has focused on automatically generating explanations\footnote{Many key works on automatic explanation interpret neurons in language or vision models directly, without SAEs \citep{bau_network_2017, hernandez_natural_2022, bills2023language, choi2024scaling}. The value proposition of SAEs is that, compared to the original neurons, SAE features can be explained with much higher fidelity.} (\citet{templeton2024scaling, oneill_disentangling_2024}, \textit{inter alia}). 
To interpret a feature $i$, a basic approach is to prompt a language model with texts that have a high value of $\mathbf{z}[i]$ against those with a low value, and ask it to identify the shared concept in the high-valued texts. 
To evaluate the quality of the resulting concept description, one can use an LM to annotate a subset of texts for the presence of the concept, and measure agreement between the concept annotations and the true feature values. 
This framework yields a quantitative score for interpretability, measuring how well a feature can be explained in natural language. 

\paragraph{Other concept-based architectures.} It is helpful to contrast the capabilities of sparse autoencoders with seemingly-similar methods like concept bottleneck models \citep{koh2020concept}. While both have neurons that fire in the presence of concepts, in concept bottleneck models these concepts are \textit{prespecified} before training the model\footnote{In the original formulation, concepts are chosen by domain experts \citep{koh2020concept}; recent work uses language models to propose concepts \citep{sun_concept_2025, ludan_interpretablebydesign_2024}. In both cases, the concepts need to be known beforehand, either by an expert or an LLM.}, while in sparse autoencoders, the concepts are learned while training the model. In other words, in the case of SAEs, researchers need not know what concepts are useful beforehand.

\begin{table*}
\centering
\begin{tabular}{>{\columncolor{gray!10}}>{\raggedright\arraybackslash}p{6.3cm}>{\raggedright\arraybackslash}p{6.8cm}}
\toprule
\textbf{Neg. Results: Acting on Known Concepts} & \textbf{Pos. Results: Discovering Unknown Concepts} \\
\midrule
\textbf{Concept detection} {\footnotesize\citep{wu2025axbench, kantamneni2025sparse}} & \textbf{Hypothesis Generation} {\footnotesize\citep{movva_sparse_2025}} \\
Is the following name a basketball player? & What concepts predict engagement on news headlines? \\
Is the following entity in New York City? & What concepts predict partisanship in Congressional speeches? \\
\midrule
\textbf{Model steering} {\footnotesize\citep{wu2025axbench}} & \textbf{Biology of LLMs} {\footnotesize\citep{lindsey2025biology}} \\
Make the LLM output more sycophantic. & What concepts does an LLM represent after writing the first line of a poem? \\
Make the LLM output discuss the Golden Gate Bridge. & What concepts does an LLM represent when performing addition? \\
\bottomrule
\end{tabular}
\vspace{0.5em}
\caption{Negative SAE results \textit{act on known concepts} whereas positive SAE results focus on \textit{discovering unknown concepts}.}
\label{table:tasks}
\end{table*}

\section{Negative results: Acting on Known Concepts}
\label{sec:negative_results}

We now survey recent negative results about SAEs, with the goal of showing that the tasks considered fall under the category of acting on known concepts. This is to be contrasted with tasks that involve discovering unknown concepts, on which positive results have been shown (Section \ref{sec:positive_results}).

Two recent papers conduct large-scale evaluations of SAEs \citep{kantamneni2025sparse, wu2025axbench}. A key finding of these papers is that SAEs underperform simple baseline methods (such as logistic regression or naive prompting). We claim that these evaluations are limited to tasks involving \textbf{acting on known concepts}. Indeed, the tasks that are studied are:
\begin{enumerate}
    \item Concept detection \citep{kantamneni2025sparse, wu2025axbench}: Identifying whether a given concept appears in a text.
    \item Model steering \citep{wu2025axbench}: Steering the outputs of a language model to contain a concept.
\end{enumerate}
These are important, widely-studied problems, and understanding how SAEs perform on them is clarifying. Notice, however, that these tasks each involve first prespecifying a concept and then acting upon it. In other words, in these tasks, concepts are inputs. We now summarize these papers' findings in greater detail.

\paragraph{Concept detection.} \citet{kantamneni2025sparse} curate 113 binary classification tasks on text data, which they use to evaluate concept detection accuracy. 
For example, one task is to determine whether a given name corresponds to a basketball player. 
Another task is to determine whether a tweet conveys happy sentiment. 
For each such task, they fit a classifier to predict the presence of the concept using \texttt{Gemma-2-9B}'s representations of the final token in each text as input.
They compare this to a classifier trained on the latent representations from a \texttt{Gemma-2-9B} SAE.
They further examine class imbalance, data scarcity, and label noise. 
In each setting, the classifier trained on top of the SAE does not outperform the classifier trained directly from the LM. 

\citet{wu2025axbench} follow a similar approach. 
Starting from a list of 500 concepts, for each concept, they generate synthetic texts that either do or do not contain the concept. 
In addition to logistic regression using \texttt{Gemma-2-2B} representations, they train several other representation-based concept detection methods.
They also include methods that do not use representations at all, such as prompting an LLM to identify whether the concept is present in the text, as well as bag-of-words. 
Four such baselines, including logistic regression and prompting, outperform the SAE.

\paragraph{Model steering.} \citet{wu2025axbench} also study model steering. 
Given a user prompt and a concept, like ``where should I visit today?'' and ``Golden Gate Bridge,'' they evaluate whether the model can generate a response that is fluent, relates to the prompt, and includes the concept. 
An LLM judge scores each attribute.
To steer with an SAE, they identify the SAE feature that is most predictive of the concept's presence, and they generate a response after increasing the value of this feature.
Non-SAE methods include editing activations with a steering vector \citep{marks_geometry_2024}, finetuning on responses containing the concept, or simply prompting to include the concept in the response.
Prompting and finetuning both outperform SAE-based steering.

\paragraph{What explains these negative results?} We suggest some reasons why SAEs underperform baselines on these tasks. 
For concept detection, recall that SAEs are trained to reconstruct the LM token representations. 
A reconstruction encodes strictly less information about a token than the original LM representation.
It follows that, compared to the original representation, there is less information available in the SAE representation to predict the presence of a concept. 
For model steering, prompting performs well because LLMs are finetuned to be adept at instruction-following, and including a concept in a response falls well within this paradigm.

The empirical results from both papers underscore an intuition that there are many natural methods besides SAEs to act on known concepts. (Though, methodological innovations may make SAEs more competitive at these tasks as well \citep{arad2025saes}.) Crucially, however, these baselines are less equipped to perform another simple task: to enumerate a list of unknown concepts that satisfy some objective. This, as we show in the next section, forms the basis for tasks on which SAEs have a comparative advantage.

\section{Positive results: Discovering Unknown Concepts}
\label{sec:positive_results}

We now describe two positive results using SAEs\footnote{Note that \citet{lindsey2025biology} use sparse coding methods that differ slightly from SAEs (see note at end of \S\ref{sec:primer}).} \citep{movva_sparse_2025, lindsey2025biology}, which focus on the following tasks:
\begin{enumerate}
    \item Hypothesis generation \citep{movva_sparse_2025}: Identifying open-ended natural language concepts that predict a target variable.
    \item Explaining language model outputs (``Biology of LLMs'') \citep{lindsey2025biology}: Describing the concepts a language model uses to perform various tasks (e.g., poem completion or addition).
\end{enumerate}

We claim that these tasks are examples of \textbf{discovering unknown concepts}. To explain this, we summarize their findings in greater detail.

\paragraph{Hypothesis generation.} \citet{movva_sparse_2025} study tasks where a large dataset of texts is annotated with a target variable, and the goal is to understand what concepts in the text predict the target. 
For example, one such dataset consists of news headlines and numerical engagement levels.
While a traditional analysis of such a dataset may be hypothesis-driven (e.g., \citet{robertson_negativity_2023} study how negativity affects engagement), here, the task is to extract concepts with no prior specification.
Such an approach can surface unknown concepts, which can be used as hypotheses for further study.

They (1) train an SAE on dense text embeddings; (2) select SAE features that predict the target; and (3) run autointerpretation to explain the selected features, which become hypotheses (i.e., ``\textit{headlines that contain \{concept\} receive more engagement}''). 
They find that the resulting hypotheses outperform those generated without an SAE, either by skipping step 1 and selecting features from text embeddings, or by using a different pipeline altogether (like LLM prompting, topic modeling, or $n$-grams). 
Compared to these baselines, it produces more statistically significant hypotheses, and raters identify these hypotheses as more helpful.

\paragraph{Mechanistic explanation of LM outputs.} \citet{lindsey2025biology} explain how language models generate text that completes a task.
For example, prompted to write a rhyming couplet, an LM generates ``\textit{He saw a carrot and had to grab it} (line 1) / \textit{His hunger was like a starving rabbit} (line 2).''
They ask: what is the internal mechanism through which the LM rhymes the end of line 2 with the end of line 1?
They find that, immediately after generating line 1, which ends in ``it,'' there is an active SAE neuron corresponding to ``\textit{words rhyming with `it'},'' as well as a neuron for ``\textit{rabbit}.''
The SAE, therefore, suggests that the LM plans line 2 immediately after generating line 1, rather than improvising a rhyming final token only after generating the first part of line 2. 
They confirm this with further intervention experiments.
In another case, they look at how a model computes ``36+59'' in natural language. 
They find active neurons for ``\textit{units digit 5},'' and ``\textit{addition problems of $\sim$40 plus $\sim$50},''  which combine to produce ``95.'' 
These specific routes of task completion are difficult to forecast, underscoring how this analysis requires discovering unknown concepts.

\paragraph{What explains these positive results?} 
In hypothesis generation, the goal is to find concepts that predict a target variable; in explaining LM behaviors, it is to find concepts that are active when completing a task.
In both cases, our hope is to discover concepts that satisfy a property of interest, out of intractably many possibilities.
SAEs produce a set of concepts that is both tractable and expressive, after which selecting a subset of concepts that are relevant to the task is straightforward.
An empirical strength of the SAE is its precise concepts.
If the \textit{rabbit} feature instead activated on all animals, it would be difficult to answer whether the model improvises or plans rhymes.

Also note that after identifying concepts of interest, it is possible to computationally validate whether they satisfy the desired property. 
In particular, once the concept has been identified, it is possible to directly annotate the data for the concept. These annotations can then be used instead of the activations. At this point, the activations are no longer needed.
It is easy to evaluate whether a hypothesized headline concept indeed correlates with engagement, or whether a hypothesized LLM addition feature is active during addition. Because of this falsifiability, even if an SAE feature is unreliable (e.g., not all headlines that contain a concept activate the corresponding feature), it is possible to catch these issues downstream.
In contrast, unreliability directly harms concept detection and steering.

\paragraph{Practical examples of concept discovery.} 
This pattern---enumerating concepts and selecting those that satisfy a criterion of interest---has already been useful in several settings beyond the two prior examples. 
\citet{tjuatja_behaviorbox_2025} and \citet{jiang_interpretable_2025} use SAEs to generate features of LM outputs, and they select features that distinguish different model versions. 
The former work finds, for example, that OLMo2-13B is better than OLMo2-7B at correctly using archaic spellings in historical texts (e.g., ``wood'' for ``would'').
\citet{movva_whats_2025} use SAEs to identify which features of LM responses predict human judgments in RLHF data, revealing unsafe and surprising preferences (e.g., high win-rates for toxic responses, and low win-rates for responses emphasizing environmental sustainability).
A similar playbook has been promising for biological foundation models \citep{simon_interplm_2025a, brixi_genome_2025}; for example, \citet{adams_mechanistic_2025} train SAEs on protein LMs and identify amino acid features that predict properties like thermostability, yielding hypotheses for how structure encodes function.

While each of these research questions (``how do LM outputs differ'', ``which protein structures are most stable'') could by studied be pre-specifying concepts and testing them, SAEs provide the opportunity to discover unknown concepts. 
Therefore, it is situations where we lack existing theories, or where we wish to expand upon them \citep{mullainathan_science_2025}, where SAEs excel.

\begin{table*}
\centering
\makebox[\textwidth]{  %
\begin{tabular}{>{\raggedright\arraybackslash}p{4cm}>{\raggedright\arraybackslash}p{11cm}}
\toprule
\textbf{Research area} & \textbf{Research problem (using SAEs to discover unknown concepts)} \\
 \midrule
 ML interpretability and explainability & Finding natural language concepts that can be used to build an inherently interpretable model. \citep{rudin2019stop} \vspace{0.2cm}\\
  & Finding natural language concepts that explain a model's predictions. \citep{lakkaraju2019faithful} \\
  \midrule
  ML fairness, bias, auditing, and safety &  In what ways do LLMs stereotype different demographic groups? \citep{lucy-bamman-2021-gender} \vspace{0.2cm}\\
  & What features are high-stakes LLM-based decision tools using? \citep{gaebler2024auditing}\vspace{0.2cm}\\
  & What undesirable behaviors do LLMs exhibit? \citep{dai2025aggregated}\\
 \midrule
Social and health sciences & How has language about immigration changed over time in Congressional speeches? \citep{card2022computational} \vspace{0.2cm}\\
 & What symptoms (recorded in medical records) predict clinical outcomes? \citep{huang2019clinicalbert} \vspace{0.2cm}\\
 & What information from court hearings do judges use when making bail decisions? \citep{zhang2024tipping}\vspace{0.2cm}\\
 & What features explain the difference in accuracy between predictive models and theory-grounded models? \citep{fudenberg2022measuring} \vspace{0.2cm}\\
 & Are ML models using illegitimate features (in the context of making a scientific claim)? \citep{kapoor2023leakage} \\
\bottomrule
\end{tabular}
}
\vspace{0.5em}
\caption{Example research problems where SAEs can be applied to discover unknown concepts.}
\vspace{-0.5em}
\label{tab:research-problems}
\end{table*}

\section{Call to Action: Future Use Cases for SAEs}
\label{sec:uses}

Having conceptualized where SAEs are useful (discovering unknown concepts), we outline research areas where such a capability can be useful. In particular, while initial excitement about SAEs was shared primarily by researchers in mechanistic interpretability \citep{sharkey2025open}, we believe that clarifying the comparative advantage of SAEs reveals a significantly broader set of uses. The use cases we outline focus on the ability of SAEs to discover unknown concepts.

Broadly, these use cases fall under two categories: using SAEs to understand language models (in ML fairness, interpretability, explainability, auditing, and safety) and using SAEs to understand the world (e.g., in the social sciences and healthcare). We summarize these potential use cases of SAEs for different research problems in Table \ref{tab:research-problems}.

\paragraph{ML fairness, interpretability, explainability, auditing, and safety.} Each of these areas aim to understand and build models with desiderata beyond accuracy in mind. Here, we see significant opportunity for SAEs. For example, SAEs can be used to identify natural language concepts that can explain black box model behavior \citep{lakkaraju2019faithful}. Then, by identifying the concepts that are used, it is possible to build models that are inherently interpretable \citep{rudin2019stop}, and that incorporate only features that we want (e.g., that are considered fair, avoid spurious correlations, etc). 

SAEs are particularly valuable for studying models with unstructured text inputs or outputs. For example, whereas existing work documents how specific demographic features affect LLM-based decision-making (e.g., in hiring \citep{gaebler2024auditing}), it is possible to use SAEs to uncover a wider range of input features that may affect LLM-generated decisions. Similarly, while demographic information has been demonstrated to affect LLM outputs in specific ways, it is possible to use SAEs to uncover a wider range of output features that are affected by variations in inputs. 

\paragraph{Health and social sciences.} 

A wide variety of disciplines (e.g., sociology, economics, healthcare) have sought to leverage large text datasets. This has led to prominent work developing and applying methods for ``text as data'' \citep{grimmer2010bayesian, gentzkow2019text}. These methods often attempt to discover interpretable patterns in text data---for example, quantifying changes in the language used to discuss immigrants \citep{garg_word_2018, card2022computational}, or identifying features of clinical notes that predict health outcomes \citep{harrigian_characterization_2023, hsu_machine_2025}. Existing methods automate these tasks through simple text features such as keywords or $n$-grams, or through topic models. These methods are limited by the expressivity of these features: topic models and keywords do not precisely capture the range of concepts present in text. In contrast, text embeddings can better capture the information present in text, but are uninterpretable. SAEs essentially convert uninterpretable text embeddings into interpretable text embeddings, enabling their use for the same applications as previous keyword or topic model methods---i.e., discovering concepts that reveal patterns in text---but potentially with significantly higher quality. SAEs provide a way to revisit important problems studied using text as data, but with the capabilities of modern language models.

Stepping back, SAEs are a promising tool for bridging the ``prediction-explanation gap.'' There are many settings in which text data have been shown to enable much greater predictive accuracy than existing human-specified features. While developing methods to quantify or improve predictive accuracy may be of independent interest, a growing line of work has suggested the need to bridge the gap between prediction and explanation \citep{hofman2017prediction, hofman2021integrating}. Traditionally, scientific disciplines have sought to \textit{explain} phenomena, rather than only predict outcomes. For example, \citet{fudenberg2022measuring} and \citet{ludwig2024machine} each show gaps between predictive accuracy of ML models that take in all available features and models that take in existing human-specified features. This gap suggests that existing theories are incomplete, leading to work that has sought to build automated approaches for closing this gap: discovering interpretable features that are predictive. SAEs are a promising tool for this task \citep{movva_sparse_2025}. SAEs can help close the prediction-explanation gap by converting black box representations into interpretable representations. These interpretable representations both capture much of the predictive power of the black box representations, while also enabling us to make predictions in terms of natural language concepts. 

For example, one important motivation for closing this gap is spurious correlation: it is well established that strong predictive performance can be misleading in ML-based science applications, underscoring the need for explanation \citep{kapoor2024reforms, messeri2024artificial, del2024prediction, 10.1214/10-STS330}. ML models with high accuracy may use illegitimate or spurious features, as has been shown repeatedly, for example, in healthcare applications \citep{ross2021epic, chiavegatto2021data, zech2018variable, gichoya2022ai, hill2024risk}. In unstructured text data, discovering these illegitimate features can be difficult. SAEs provide one way of discovering these features. This capability extends past methods that \textit{prespecify} concepts to be used for prediction with unstructured data \citep{koh2020concept}.

\section{Alternative Views}
There has been considerable debate about the usefulness of sparse autoencoders, and the degree to which they are a fruitful research direction (e.g., \citet{deepmind2024sae}). Those who take the negative position on SAEs have pointed to negative results showing that SAEs underperform baselines \citep{wu2025axbench, kantamneni2025sparse}. The positive position we take is based on our conceptual framework, which establishes that these negative results lie in the regime of \textit{acting on known concepts}, while there exist exciting positive results in the regime of \textit{discovering unknown concepts}.

While we have argued that this suggests promise in further research using SAEs to discover unknowns, an alternate view is that the most important problems fall under the category of acting on knowns (e.g., language model steering). Researchers who take this perspective would therefore argue that our analysis does not imply that more attention should be devoted to SAE research. However, as we have laid out in \Cref{sec:uses}, we believe there are a diverse set of important research questions---spanning multiple fields---in the regime of discovering unknown concepts.

On the other hand, other researchers may hold the view that our position is too strong in suggesting that the known-unknown dichotomy explains negative and positive results. While we do show that negative results fall in the known concepts regime---and that there are basic reasons to suggest why SAEs may be less successful in these regimes---we do not intend to rule out the possibility that SAEs will also be useful in such a setting. More research in this direction is likely to be useful. Researchers who hold this position would in any case, however, agree that despite existing negative results, SAE research holds promise.

\section{Conclusion}

In this position paper, we presented a conceptual framework for understanding different applications of SAEs. The framework separates uses that act on known concepts (like steering or probing) from uses that act on unknown concepts (like hypothesis generation or LLM biology). We illustrate this by presenting several case studies in \Cref{sec:negative_results} and \Cref{sec:positive_results}. The distinction reconciles tensions between negative results on SAEs and positive results. Overall, the distinction helps focus future research: while SAEs may be generally useful (especially as methods improve), SAEs are an especially promising tool for discovering unknowns. We discuss these use cases in \Cref{sec:uses}. Overall, our paper attempts to better understand the different uses of SAEs. By clarifying distinctions between these uses, we hope that future work can pursue the directions of promise, further study these distinctions, and design SAE-based methods with a clearer understanding of what the end goals are.

{
\raggedright
\bibliographystyle{alpha}
\bibliography{bib,zotero}

\begin{thebibliography}{68}
\providecommand{\natexlab}[1]{#1}
\providecommand{\url}[1]{\texttt{#1}}
\expandafter\ifx\csname urlstyle\endcsname\relax
  \providecommand{\doi}[1]{doi: #1}\else
  \providecommand{\doi}{doi: \begingroup \urlstyle{rm}\Url}\fi

\bibitem[Adams et~al.(2025)Adams, Bai, Lee, Yu, and AlQuraishi]{adams_mechanistic_2025}
E.~Adams, L.~Bai, M.~Lee, Y.~Yu, and M.~AlQuraishi.
\newblock From {{Mechanistic Interpretability}} to {{Mechanistic Biology}}: {{Training}}, {{Evaluating}}, and {{Interpreting Sparse Autoencoders}} on {{Protein Language Models}}, Feb. 2025.

\bibitem[Anthropic(2024)]{anthropic2024goldengate}
Anthropic.
\newblock {Golden Gate Claude}, 2024.

\bibitem[Arad et~al.(2025)Arad, Mueller, and Belinkov]{arad2025saes}
D.~Arad, A.~Mueller, and Y.~Belinkov.
\newblock {SAE}s are good for steering--if you select the right features.
\newblock \emph{arXiv preprint arXiv:2505.20063}, 2025.

\bibitem[Baldi and Hornik(1989)]{baldi_neural_1989}
P.~Baldi and K.~Hornik.
\newblock Neural networks and principal component analysis: {{Learning}} from examples without local minima.
\newblock \emph{Neural Networks}, 2\penalty0 (1):\penalty0 53--58, Jan. 1989.
\newblock ISSN 0893-6080.

\bibitem[Bau et~al.(2017)Bau, Zhou, Khosla, Oliva, and Torralba]{bau_network_2017}
D.~Bau, B.~Zhou, A.~Khosla, A.~Oliva, and A.~Torralba.
\newblock Network {{Dissection}}: {{Quantifying Interpretability}} of {{Deep Visual Representations}}, Apr. 2017.

\bibitem[Belinkov(2022)]{belinkov_probing_2022}
Y.~Belinkov.
\newblock Probing {{Classifiers}}: {{Promises}}, {{Shortcomings}}, and {{Advances}}.
\newblock \emph{Computational Linguistics}, 48\penalty0 (1):\penalty0 207--219, Apr. 2022.
\newblock ISSN 0891-2017.

\bibitem[Bills et~al.(2023)Bills, Cammarata, Mossing, Tillman, Gao, Goh, Sutskever, Leike, Wu, and Saunders]{bills2023language}
S.~Bills, N.~Cammarata, D.~Mossing, H.~Tillman, L.~Gao, G.~Goh, I.~Sutskever, J.~Leike, J.~Wu, and W.~Saunders.
\newblock Language models can explain neurons in language models.
\newblock \url{https://openaipublic.blob.core.windows.net/neuron-explainer/paper/index.html}, 2023.

\bibitem[Bricken et~al.(2023)Bricken, Templeton, Batson, Chen, Jermyn, Conerly, Turner, Anil, Denison, Askell, et~al.]{bricken2023towards}
T.~Bricken, A.~Templeton, J.~Batson, B.~Chen, A.~Jermyn, T.~Conerly, N.~Turner, C.~Anil, C.~Denison, A.~Askell, et~al.
\newblock Towards monosemanticity: Decomposing language models with dictionary learning.
\newblock \emph{Transformer Circuits Thread}, 2, 2023.

\bibitem[Brixi et~al.(2025)Brixi, Durrant, Ku, Poli, Brockman, Chang, Gonzalez, King, Li, Merchant, Naghipourfar, Nguyen, {Ricci-Tam}, Romero, Sun, Taghibakshi, Vorontsov, Yang, Deng, Gorton, Nguyen, Wang, Adams, Baccus, Dillmann, Ermon, Guo, Ilango, Janik, Lu, Mehta, Mofrad, Ng, Pannu, R{\'e}, Schmok, John, Sullivan, Zhu, Zynda, Balsam, Collison, Costa, {Hernandez-Boussard}, Ho, Liu, McGrath, Powell, Burke, Goodarzi, Hsu, and Hie]{brixi_genome_2025}
G.~Brixi, M.~G. Durrant, J.~Ku, M.~Poli, G.~Brockman, D.~Chang, G.~A. Gonzalez, S.~H. King, D.~B. Li, A.~T. Merchant, M.~Naghipourfar, E.~Nguyen, C.~{Ricci-Tam}, D.~W. Romero, G.~Sun, A.~Taghibakshi, A.~Vorontsov, B.~Yang, M.~Deng, L.~Gorton, N.~Nguyen, N.~K. Wang, E.~Adams, S.~A. Baccus, S.~Dillmann, S.~Ermon, D.~Guo, R.~Ilango, K.~Janik, A.~X. Lu, R.~Mehta, M.~R.~K. Mofrad, M.~Y. Ng, J.~Pannu, C.~R{\'e}, J.~C. Schmok, J.~S. John, J.~Sullivan, K.~Zhu, G.~Zynda, D.~Balsam, P.~Collison, A.~B. Costa, T.~{Hernandez-Boussard}, E.~Ho, M.-Y. Liu, T.~McGrath, K.~Powell, D.~P. Burke, H.~Goodarzi, P.~D. Hsu, and B.~L. Hie.
\newblock Genome modeling and design across all domains of life with {{Evo}} 2, Feb. 2025.

\bibitem[Bussmann et~al.(2025)Bussmann, Nabeshima, Karvonen, and Nanda]{bussmann_learning_2025}
B.~Bussmann, N.~Nabeshima, A.~Karvonen, and N.~Nanda.
\newblock Learning {{Multi-Level Features}} with {{Matryoshka Sparse Autoencoders}}, Mar. 2025.

\bibitem[Card et~al.(2022)Card, Chang, Becker, Mendelsohn, Voigt, Boustan, Abramitzky, and Jurafsky]{card2022computational}
D.~Card, S.~Chang, C.~Becker, J.~Mendelsohn, R.~Voigt, L.~Boustan, R.~Abramitzky, and D.~Jurafsky.
\newblock Computational analysis of 140 years of {US} political speeches reveals more positive but increasingly polarized framing of immigration.
\newblock \emph{Proceedings of the National Academy of Sciences}, 119\penalty0 (31):\penalty0 e2120510119, 2022.

\bibitem[Chiavegatto~Filho et~al.(2021)Chiavegatto~Filho, Batista, and Dos~Santos]{chiavegatto2021data}
A.~Chiavegatto~Filho, A.~F. D.~M. Batista, and H.~G. Dos~Santos.
\newblock {Data leakage in health outcomes prediction with machine learning. Comment on “Prediction of incident hypertension within the next year: prospective study using statewide electronic health records and machine learning”}.
\newblock \emph{Journal of medical Internet research}, 23\penalty0 (2):\penalty0 e10969, 2021.

\bibitem[Choi et~al.(2024)Choi, Huang, Meng, Johnson, Steinhardt, and Schwettmann]{choi2024scaling}
D.~Choi, V.~Huang, K.~Meng, D.~D. Johnson, J.~Steinhardt, and S.~Schwettmann.
\newblock Scaling automatic neuron description, October 2024.
\newblock URL \url{https://transluce.org/neuron-descriptions}.
\newblock Published online at Transluce AI.

\bibitem[Coates and Ng(2011)]{coates_importance_2011}
A.~Coates and A.~Y. Ng.
\newblock The importance of encoding versus training with sparse coding and vector quantization.
\newblock In \emph{Proceedings of the 28th {{International Conference}} on {{International Conference}} on {{Machine Learning}}}, {{ICML}}'11, pages 921--928. Omnipress, June 2011.
\newblock ISBN 978-1-4503-0619-5.

\bibitem[Cunningham et~al.(2023)Cunningham, Ewart, Riggs, Huben, and Sharkey]{cunningham_sparse_2023}
H.~Cunningham, A.~Ewart, L.~Riggs, R.~Huben, and L.~Sharkey.
\newblock Sparse {{Autoencoders Find Highly Interpretable Features}} in {{Language Models}}, Oct. 2023.

\bibitem[Cywi{\'n}ski and Deja(2025)]{cywinski2025saeuron}
B.~Cywi{\'n}ski and K.~Deja.
\newblock Saeuron: Interpretable concept unlearning in diffusion models with sparse autoencoders.
\newblock \emph{arXiv preprint arXiv:2501.18052}, 2025.

\bibitem[Dai et~al.(2025)Dai, Raji, Recht, and Chen]{dai2025aggregated}
J.~Dai, I.~D. Raji, B.~Recht, and I.~Y. Chen.
\newblock Aggregated individual reporting for post-deployment evaluation.
\newblock \emph{arXiv preprint arXiv:2506.18133}, 2025.

\bibitem[Del~Giudice et~al.(2024)]{del2024prediction}
M.~Del~Giudice et~al.
\newblock The prediction-explanation fallacy: A pervasive problem in scientific applications of machine learning.
\newblock \emph{Methodology}, 20\penalty0 (1):\penalty0 22--46, 2024.

\bibitem[Elhage et~al.(2022)Elhage, Hume, Olsson, Schiefer, Henighan, Kravec, {Hatfield-Dodds}, Lasenby, Drain, Chen, Grosse, McCandlish, Kaplan, Amodei, Wattenberg, and Olah]{elhage_toy_2022}
N.~Elhage, T.~Hume, C.~Olsson, N.~Schiefer, T.~Henighan, S.~Kravec, Z.~{Hatfield-Dodds}, R.~Lasenby, D.~Drain, C.~Chen, R.~Grosse, S.~McCandlish, J.~Kaplan, D.~Amodei, M.~Wattenberg, and C.~Olah.
\newblock Toy {{Models}} of {{Superposition}}, Sept. 2022.
\newblock Comment: Also available at https://transformer-circuits.pub/2022/toy\_model/index.html.

\bibitem[Farrell et~al.(2024)Farrell, Lau, and Conmy]{farrell2024applying}
E.~Farrell, Y.-T. Lau, and A.~Conmy.
\newblock Applying sparse autoencoders to unlearn knowledge in language models.
\newblock \emph{arXiv preprint arXiv:2410.19278}, 2024.

\bibitem[Fudenberg et~al.(2022)Fudenberg, Kleinberg, Liang, and Mullainathan]{fudenberg2022measuring}
D.~Fudenberg, J.~Kleinberg, A.~Liang, and S.~Mullainathan.
\newblock Measuring the completeness of economic models.
\newblock \emph{Journal of Political Economy}, 130\penalty0 (4):\penalty0 956--990, 2022.

\bibitem[Gaebler et~al.(2024)Gaebler, Goel, Huq, and Tambe]{gaebler2024auditing}
J.~D. Gaebler, S.~Goel, A.~Huq, and P.~Tambe.
\newblock Auditing large language models for race \& gender disparities: Implications for artificial intelligence-based hiring.
\newblock \emph{Behavioral Science \& Policy}, 10\penalty0 (2):\penalty0 46--55, 2024.

\bibitem[Gao et~al.(2024)Gao, la~Tour, Tillman, Goh, Troll, Radford, Sutskever, Leike, and Wu]{gao_scaling_2024}
L.~Gao, T.~D. la~Tour, H.~Tillman, G.~Goh, R.~Troll, A.~Radford, I.~Sutskever, J.~Leike, and J.~Wu.
\newblock Scaling and evaluating sparse autoencoders, June 2024.

\bibitem[Garg et~al.(2018)Garg, Schiebinger, Jurafsky, and Zou]{garg_word_2018}
N.~Garg, L.~Schiebinger, D.~Jurafsky, and J.~Zou.
\newblock Word embeddings quantify 100 years of gender and ethnic stereotypes.
\newblock \emph{Proceedings of the National Academy of Sciences}, 115\penalty0 (16):\penalty0 E3635--E3644, Apr. 2018.

\bibitem[Gentzkow et~al.(2019)Gentzkow, Kelly, and Taddy]{gentzkow2019text}
M.~Gentzkow, B.~Kelly, and M.~Taddy.
\newblock Text as data.
\newblock \emph{Journal of Economic Literature}, 57\penalty0 (3):\penalty0 535--574, 2019.

\bibitem[Gichoya et~al.(2022)Gichoya, Banerjee, Bhimireddy, Burns, Celi, Chen, Correa, Dullerud, Ghassemi, Huang, et~al.]{gichoya2022ai}
J.~W. Gichoya, I.~Banerjee, A.~R. Bhimireddy, J.~L. Burns, L.~A. Celi, L.-C. Chen, R.~Correa, N.~Dullerud, M.~Ghassemi, S.-C. Huang, et~al.
\newblock Ai recognition of patient race in medical imaging: a modelling study.
\newblock \emph{The Lancet Digital Health}, 4\penalty0 (6):\penalty0 e406--e414, 2022.

\bibitem[Grimmer(2010)]{grimmer2010bayesian}
J.~Grimmer.
\newblock A bayesian hierarchical topic model for political texts: Measuring expressed agendas in senate press releases.
\newblock \emph{Political analysis}, 18\penalty0 (1):\penalty0 1--35, 2010.

\bibitem[Harrigian et~al.(2023)Harrigian, Zirikly, Chee, Ahmad, Links, Saha, Beach, and Dredze]{harrigian_characterization_2023}
K.~Harrigian, A.~Zirikly, B.~Chee, A.~Ahmad, A.~Links, S.~Saha, M.~C. Beach, and M.~Dredze.
\newblock Characterization of {{Stigmatizing Language}} in {{Medical Records}}.
\newblock In A.~Rogers, J.~{Boyd-Graber}, and N.~Okazaki, editors, \emph{Proceedings of the 61st {{Annual Meeting}} of the {{Association}} for {{Computational Linguistics}} ({{Volume}} 2: {{Short Papers}})}, pages 312--329. Association for Computational Linguistics, July 2023.

\bibitem[Hernandez et~al.(2022)Hernandez, Schwettmann, Bau, Bagashvili, Torralba, and Andreas]{hernandez_natural_2022}
E.~Hernandez, S.~Schwettmann, D.~Bau, T.~Bagashvili, A.~Torralba, and J.~Andreas.
\newblock Natural {{Language Descriptions}} of {{Deep Visual Features}}, Apr. 2022.

\bibitem[Hill et~al.(2024)Hill, Koback, and Schilling]{hill2024risk}
B.~G. Hill, F.~L. Koback, and P.~L. Schilling.
\newblock The risk of shortcutting in deep learning algorithms for medical imaging research.
\newblock \emph{Scientific Reports}, 14\penalty0 (1):\penalty0 29224, 2024.

\bibitem[Hinton and Salakhutdinov(2006)]{hinton_reducing_2006}
G.~E. Hinton and R.~R. Salakhutdinov.
\newblock Reducing the {{Dimensionality}} of {{Data}} with {{Neural Networks}}.
\newblock \emph{Science}, 313\penalty0 (5786):\penalty0 504--507, July 2006.

\bibitem[Hofman et~al.(2017)Hofman, Sharma, and Watts]{hofman2017prediction}
J.~M. Hofman, A.~Sharma, and D.~J. Watts.
\newblock Prediction and explanation in social systems.
\newblock \emph{Science}, 355\penalty0 (6324):\penalty0 486--488, 2017.

\bibitem[Hofman et~al.(2021)Hofman, Watts, Athey, Garip, Griffiths, Kleinberg, Margetts, Mullainathan, Salganik, Vazire, et~al.]{hofman2021integrating}
J.~M. Hofman, D.~J. Watts, S.~Athey, F.~Garip, T.~L. Griffiths, J.~Kleinberg, H.~Margetts, S.~Mullainathan, M.~J. Salganik, S.~Vazire, et~al.
\newblock Integrating explanation and prediction in computational social science.
\newblock \emph{Nature}, 595\penalty0 (7866):\penalty0 181--188, 2021.

\bibitem[Hsu et~al.(2025)Hsu, Obermeyer, and Tan]{hsu_machine_2025}
C.-C. Hsu, Z.~Obermeyer, and C.~Tan.
\newblock A machine learning model using clinical notes to identify physician fatigue.
\newblock \emph{Nature Communications}, 16\penalty0 (1):\penalty0 5791, July 2025.
\newblock ISSN 2041-1723.

\bibitem[Huang et~al.(2019)Huang, Altosaar, and Ranganath]{huang2019clinicalbert}
K.~Huang, J.~Altosaar, and R.~Ranganath.
\newblock Clinicalbert: Modeling clinical notes and predicting hospital readmission.
\newblock \emph{arXiv preprint arXiv:1904.05342}, 2019.

\bibitem[Jiang et~al.(2025)Jiang, Sun, Dunlap, Smith, and Nanda]{jiang_interpretable_2025}
N.~Jiang, X.~Sun, L.~Dunlap, L.~Smith, and N.~Nanda.
\newblock Interpretable {{Embeddings}} with {{Sparse Autoencoders}}: {{A Data Analysis Toolkit}}, Dec. 2025.

\bibitem[Kantamneni et~al.(2025)Kantamneni, Engels, Rajamanoharan, Tegmark, and Nanda]{kantamneni2025sparse}
S.~Kantamneni, J.~Engels, S.~Rajamanoharan, M.~Tegmark, and N.~Nanda.
\newblock Are sparse autoencoders useful? a case study in sparse probing.
\newblock \emph{arXiv preprint arXiv:2502.16681}, 2025.

\bibitem[Kapoor and Narayanan(2023)]{kapoor2023leakage}
S.~Kapoor and A.~Narayanan.
\newblock Leakage and the reproducibility crisis in machine-learning-based science.
\newblock \emph{Patterns}, 4\penalty0 (9), 2023.

\bibitem[Kapoor et~al.(2024)Kapoor, Cantrell, Peng, Pham, Bail, Gundersen, Hofman, Hullman, Lones, Malik, et~al.]{kapoor2024reforms}
S.~Kapoor, E.~M. Cantrell, K.~Peng, T.~H. Pham, C.~A. Bail, O.~E. Gundersen, J.~M. Hofman, J.~Hullman, M.~A. Lones, M.~M. Malik, et~al.
\newblock Reforms: Consensus-based recommendations for machine-learning-based science.
\newblock \emph{Science Advances}, 10\penalty0 (18):\penalty0 eadk3452, 2024.

\bibitem[Koh et~al.(2020)Koh, Nguyen, Tang, Mussmann, Pierson, Kim, and Liang]{koh2020concept}
P.~W. Koh, T.~Nguyen, Y.~S. Tang, S.~Mussmann, E.~Pierson, B.~Kim, and P.~Liang.
\newblock Concept bottleneck models.
\newblock In \emph{International conference on machine learning}, pages 5338--5348. PMLR, 2020.

\bibitem[Lakkaraju et~al.(2019)Lakkaraju, Kamar, Caruana, and Leskovec]{lakkaraju2019faithful}
H.~Lakkaraju, E.~Kamar, R.~Caruana, and J.~Leskovec.
\newblock Faithful and customizable explanations of black box models.
\newblock In \emph{Proceedings of the 2019 AAAI/ACM Conference on AI, Ethics, and Society}, pages 131--138, 2019.

\bibitem[Lindsey et~al.(2025)Lindsey, Gurnee, Ameisen, Chen, Pearce, Turner, Citro, Abrahams, Carter, Hosmer, Marcus, Sklar, Templeton, Bricken, McDougall, Cunningham, Henighan, Jermyn, Jones, Persic, Qi, Thompson, Zimmerman, Rivoire, Conerly, Olah, and Batson]{lindsey2025biology}
J.~Lindsey, W.~Gurnee, E.~Ameisen, B.~Chen, A.~Pearce, N.~L. Turner, C.~Citro, D.~Abrahams, S.~Carter, B.~Hosmer, J.~Marcus, M.~Sklar, A.~Templeton, T.~Bricken, C.~McDougall, H.~Cunningham, T.~Henighan, A.~Jermyn, A.~Jones, A.~Persic, Z.~Qi, T.~B. Thompson, S.~Zimmerman, K.~Rivoire, T.~Conerly, C.~Olah, and J.~Batson.
\newblock On the biology of a large language model.
\newblock \emph{Transformer Circuits Thread}, 2025.

\bibitem[Liu et~al.(2019)Liu, Gardner, Belinkov, Peters, and Smith]{liu_linguistic_2019}
N.~F. Liu, M.~Gardner, Y.~Belinkov, M.~E. Peters, and N.~A. Smith.
\newblock Linguistic {{Knowledge}} and {{Transferability}} of {{Contextual Representations}}, Apr. 2019.

\bibitem[Lucy and Bamman(2021)]{lucy-bamman-2021-gender}
L.~Lucy and D.~Bamman.
\newblock Gender and representation bias in {GPT}-3 generated stories.
\newblock In N.~Akoury, F.~Brahman, S.~Chaturvedi, E.~Clark, M.~Iyyer, and L.~J. Martin, editors, \emph{Proceedings of the Third Workshop on Narrative Understanding}, pages 48--55, Virtual, June 2021. Association for Computational Linguistics.
\newblock \doi{10.18653/v1/2021.nuse-1.5}.
\newblock URL \url{https://aclanthology.org/2021.nuse-1.5/}.

\bibitem[Ludan et~al.(2024)Ludan, Lyu, Yang, Dugan, Yatskar, and {Callison-Burch}]{ludan_interpretablebydesign_2024}
J.~M. Ludan, Q.~Lyu, Y.~Yang, L.~Dugan, M.~Yatskar, and C.~{Callison-Burch}.
\newblock Interpretable-by-{{Design Text Understanding}} with {{Iteratively Generated Concept Bottleneck}}, Apr. 2024.

\bibitem[Ludwig and Mullainathan(2024)]{ludwig2024machine}
J.~Ludwig and S.~Mullainathan.
\newblock Machine learning as a tool for hypothesis generation.
\newblock \emph{The Quarterly Journal of Economics}, 139\penalty0 (2):\penalty0 751--827, 2024.

\bibitem[Makhzani and Frey(2014)]{makhzani_ksparse_2014}
A.~Makhzani and B.~Frey.
\newblock K-{{Sparse Autoencoders}}, Mar. 2014.

\bibitem[Marks and Tegmark(2024)]{marks_geometry_2024}
S.~Marks and M.~Tegmark.
\newblock The {{Geometry}} of {{Truth}}: {{Emergent Linear Structure}} in {{Large Language Model Representations}} of {{True}}/{{False Datasets}}, Aug. 2024.

\bibitem[Messeri and Crockett(2024)]{messeri2024artificial}
L.~Messeri and M.~Crockett.
\newblock Artificial intelligence and illusions of understanding in scientific research.
\newblock \emph{Nature}, 627\penalty0 (8002):\penalty0 49--58, 2024.

\bibitem[Movva et~al.(2025{\natexlab{a}})Movva, Milli, Min, and Pierson]{movva_whats_2025}
R.~Movva, S.~Milli, S.~Min, and E.~Pierson.
\newblock What's {{In My Human Feedback}}? {{Learning Interpretable Descriptions}} of {{Preference Data}}, Oct. 2025{\natexlab{a}}.

\bibitem[Movva et~al.(2025{\natexlab{b}})Movva, Peng, Garg, Kleinberg, and Pierson]{movva_sparse_2025}
R.~Movva, K.~Peng, N.~Garg, J.~Kleinberg, and E.~Pierson.
\newblock Sparse autoencoders for hypothesis generation.
\newblock In \emph{International Conference on Machine Learning}, pages 44997--45023. PMLR, 2025{\natexlab{b}}.

\bibitem[Mullainathan and Rambachan(2025)]{mullainathan_science_2025}
S.~Mullainathan and A.~Rambachan.
\newblock Science in the {{Age}} of {{Algorithms}}.
\newblock In \emph{The {{Economics}} of {{Transformative AI}}}. University of Chicago Press, Oct. 2025.

\bibitem[Nguyen et~al.(2025)Nguyen, Deng, Gala, Naruse, Virgo, Byun, Hazra, Gorton, Balsam, McGrath, Takei, and Kaji]{nguyen2025deploying}
N.~Nguyen, M.~Deng, D.~Gala, K.~Naruse, F.~G. Virgo, M.~Byun, D.~Hazra, L.~Gorton, D.~Balsam, T.~McGrath, M.~Takei, and Y.~Kaji.
\newblock Deploying interpretability to production with rakuten: Sae probes for pii detection.
\newblock \emph{Goodfire}, 2025.
\newblock https://www.goodfire.ai/blog/deploying-interpretability-to-production-with-rakuten.

\bibitem[O'Neill et~al.(2024)O'Neill, Ye, Iyer, and Wu]{oneill_disentangling_2024}
C.~O'Neill, C.~Ye, K.~Iyer, and J.~F. Wu.
\newblock Disentangling {{Dense Embeddings}} with {{Sparse Autoencoders}}, Aug. 2024.

\bibitem[Paulo et~al.(2025)Paulo, Shabalin, and Belrose]{paulo_transcoders_2025}
G.~Paulo, S.~Shabalin, and N.~Belrose.
\newblock Transcoders {{Beat Sparse Autoencoders}} for {{Interpretability}}, Feb. 2025.

\bibitem[Robertson et~al.(2023)Robertson, Pr{\"o}llochs, Schwarzenegger, P{\"a}rnamets, Van~Bavel, and Feuerriegel]{robertson_negativity_2023}
C.~E. Robertson, N.~Pr{\"o}llochs, K.~Schwarzenegger, P.~P{\"a}rnamets, J.~J. Van~Bavel, and S.~Feuerriegel.
\newblock Negativity drives online news consumption.
\newblock \emph{Nature Human Behaviour}, 7\penalty0 (5):\penalty0 812--822, May 2023.
\newblock ISSN 2397-3374.

\bibitem[Ross(2021)]{ross2021epic}
C.~Ross.
\newblock Epic’s sepsis algorithm is going off the rails in the real world. the use of these variables may explain why.
\newblock \emph{Stat, September}, 27, 2021.

\bibitem[Rudin(2019)]{rudin2019stop}
C.~Rudin.
\newblock Stop explaining black box machine learning models for high stakes decisions and use interpretable models instead.
\newblock \emph{Nature machine intelligence}, 1\penalty0 (5):\penalty0 206--215, 2019.

\bibitem[Sharkey et~al.(2025)Sharkey, Chughtai, Batson, Lindsey, Wu, Bushnaq, Goldowsky-Dill, Heimersheim, Ortega, Bloom, et~al.]{sharkey2025open}
L.~Sharkey, B.~Chughtai, J.~Batson, J.~Lindsey, J.~Wu, L.~Bushnaq, N.~Goldowsky-Dill, S.~Heimersheim, A.~Ortega, J.~Bloom, et~al.
\newblock Open problems in mechanistic interpretability.
\newblock \emph{arXiv preprint arXiv:2501.16496}, 2025.

\bibitem[Shmueli(2010)]{10.1214/10-STS330}
G.~Shmueli.
\newblock {To Explain or to Predict?}
\newblock \emph{Statistical Science}, 25\penalty0 (3):\penalty0 289 -- 310, 2010.
\newblock \doi{10.1214/10-STS330}.
\newblock URL \url{https://doi.org/10.1214/10-STS330}.

\bibitem[Simon and Zou(2025)]{simon_interplm_2025a}
E.~Simon and J.~Zou.
\newblock {{InterPLM}}: Discovering interpretable features in protein language models via sparse autoencoders.
\newblock \emph{Nature Methods}, 22\penalty0 (10):\penalty0 2107--2117, Oct. 2025.
\newblock ISSN 1548-7105.

\bibitem[Smith et~al.(2025)Smith, Rajamanoharan, Conmy, McDougall, Kramar, Lieberum, Shah, and Nanda]{deepmind2024sae}
L.~Smith, S.~Rajamanoharan, A.~Conmy, C.~McDougall, J.~Kramar, T.~Lieberum, R.~Shah, and N.~Nanda.
\newblock Negative results for sparse autoencoders on downstream tasks and deprioritising sae research (mechanistic interpretability team progress update), 2025.

\bibitem[Sun et~al.(2025)Sun, Oikarinen, Ustun, and Weng]{sun_concept_2025}
C.-E. Sun, T.~Oikarinen, B.~Ustun, and T.-W. Weng.
\newblock Concept {{Bottleneck Large Language Models}}, Sept. 2025.

\bibitem[Templeton et~al.(2024)Templeton, Conerly, Marcus, Lindsey, Bricken, Chen, Pearce, Citro, Ameisen, Jones, et~al.]{templeton2024scaling}
A.~Templeton, T.~Conerly, J.~Marcus, J.~Lindsey, T.~Bricken, B.~Chen, A.~Pearce, C.~Citro, E.~Ameisen, A.~Jones, et~al.
\newblock Scaling monosemanticity: Extracting interpretable features from claude 3 sonnet. transformer circuits thread, 2024.

\bibitem[Tjuatja and Neubig(2025)]{tjuatja_behaviorbox_2025}
L.~Tjuatja and G.~Neubig.
\newblock {{BehaviorBox}}: {{Automated Discovery}} of {{Fine-Grained Performance Differences Between Language Models}}, June 2025.

\bibitem[Wu et~al.(2025)Wu, Arora, Geiger, Wang, Huang, Jurafsky, Manning, and Potts]{wu2025axbench}
Z.~Wu, A.~Arora, A.~Geiger, Z.~Wang, J.~Huang, D.~Jurafsky, C.~D. Manning, and C.~Potts.
\newblock Axbench: Steering llms? even simple baselines outperform sparse autoencoders.
\newblock \emph{arXiv preprint arXiv:2501.17148}, 2025.

\bibitem[Zech et~al.(2018)Zech, Badgeley, Liu, Costa, Titano, and Oermann]{zech2018variable}
J.~R. Zech, M.~A. Badgeley, M.~Liu, A.~B. Costa, J.~J. Titano, and E.~K. Oermann.
\newblock Variable generalization performance of a deep learning model to detect pneumonia in chest radiographs: a cross-sectional study.
\newblock \emph{PLoS medicine}, 15\penalty0 (11):\penalty0 e1002683, 2018.

\bibitem[Zhang(2024)]{zhang2024tipping}
S.~Zhang.
\newblock Tipping the balance: Predictive algorithms and institutional decision-making in context.
\newblock 2024.

\end{thebibliography}
}

\end{document}